\pdfoutput=1

\documentclass[11pt]{article}

\usepackage[]{EACL2023}

\usepackage{times}
\usepackage{latexsym}

\usepackage[T1]{fontenc}

\usepackage{microtype}
\usepackage{algorithm}
\usepackage{algorithmic}
\usepackage{booktabs} 
\usepackage{enumitem}
\usepackage{bbm}
\usepackage{csquotes}
\usepackage{multirow}
\usepackage{multicol}
\usepackage{comment}
\usepackage{amsmath}
\usepackage{xcolor}
\usepackage{inconsolata}

\usepackage{times}
\usepackage{latexsym}
\usepackage{array}

\definecolor{calpolypomonagreen}{rgb}{0.12, 0.3, 0.17}
\definecolor{ao(english)}{rgb}{0.0, 0.5, 0.0}
\definecolor{cadmiumred}{rgb}{0.89, 0.0, 0.13}

\usepackage{multirow}
\usepackage{multicol}
\usepackage{amssymb}
\usepackage{pifont}
\usepackage{booktabs}
\usepackage{graphics}
\usepackage{graphicx}
\usepackage{tablefootnote}
\usepackage{footnote}
\usepackage[normalem]{ulem}
\makesavenoteenv{tabular}
\makesavenoteenv{table}

\defcitealias{rajaby-faghihi-kordjamshidi-2021-time}{(Faghini et al., 2021)}
\defcitealias{gupta-durrett-2019-tracking}{(Gupta et al., 2019)}
\defcitealias{DBLP:conf/iclr/LoshchilovH19}{(Loshchilov et al., 2019)}

\defcitealias{10.1145/3442381.3450126}{(Zhang et al., 2021)}

\newcommand\real{\textsc{Real}}
\newcommand\tslm{\textsc{TSLM}}
\newcommand\ien{\textsc{IEN}}
\newcommand\lemon{\textsc{LEMon}}
\newcommand\koala{\textsc{KoaLa}}
\newcommand\ncet{\textsc{NCET}}
\newcommand\dynapro{\textsc{DynaPro}}

\newcommand\cgli{\textsc{CGLI}}

\newcommand\ours{\textsc{MeeT}}

\newcolumntype{C}[1]{>{\PreserveBackslash\centering}p{#1}}
\newcolumntype{R}[1]{>{\PreserveBackslash\raggedleft}p{#1}}
\newcolumntype{L}[1]{>{\PreserveBackslash\raggedright}p{#1}}

\title{Entity Tracking via Effective Use of Multi-Task Learning Model and \\ Mention-guided Decoding}

\author{Janvijay Singh$^{\clubsuit}$ Fan Bai$^{\clubsuit}$  Zhen Wang$^{\diamondsuit}$ \\
  $\clubsuit$ {School of Interactive Computing, Georgia Institute of Technology} \\
  $\diamondsuit$ {Department of Computer Science and Engineering, The Ohio State University} \\
 \texttt{iamjanvijay@gatech.edu} \\
 \texttt{fan.bai@cc.gatech.edu} \\
 \texttt{wang.9215@osu.edu} \\}

\begin{document}
\maketitle
\begin{abstract}


Cross-task knowledge transfer via multi-task learning has recently made remarkable progress in general NLP tasks. 
However, entity tracking on the procedural text has not benefited from such knowledge transfer because of its distinct formulation, i.e., tracking the event flow while following structural constraints. 
State-of-the-art entity tracking approaches either design complicated model architectures or rely on task-specific pre-training to achieve good results.  
To this end, we propose \textbf{\ours{}}, a \textbf{M}ulti-task learning-\textbf{e}nabled \textbf{e}ntity \textbf{T}racking approach, which utilizes knowledge gained from general domain tasks to improve entity tracking. 
Specifically, \ours{} first fine-tunes T5, a pre-trained multi-task learning model, with entity tracking-specialized QA formats, and then employs our customized decoding strategy to satisfy the structural constraints.
\ours{} achieves state-of-the-art performances on two popular entity tracking datasets, even though it does not require any task-specific architecture design or pre-training.\footnote{Our code and data are available at \url{https://github.com/iamjanvijay/MeeT}.}



\end{abstract}

\section{Introduction}
\label{sec:intro}

\begin{figure}[!t]
\begin{center}
  \includegraphics[width=0.45\textwidth]
  {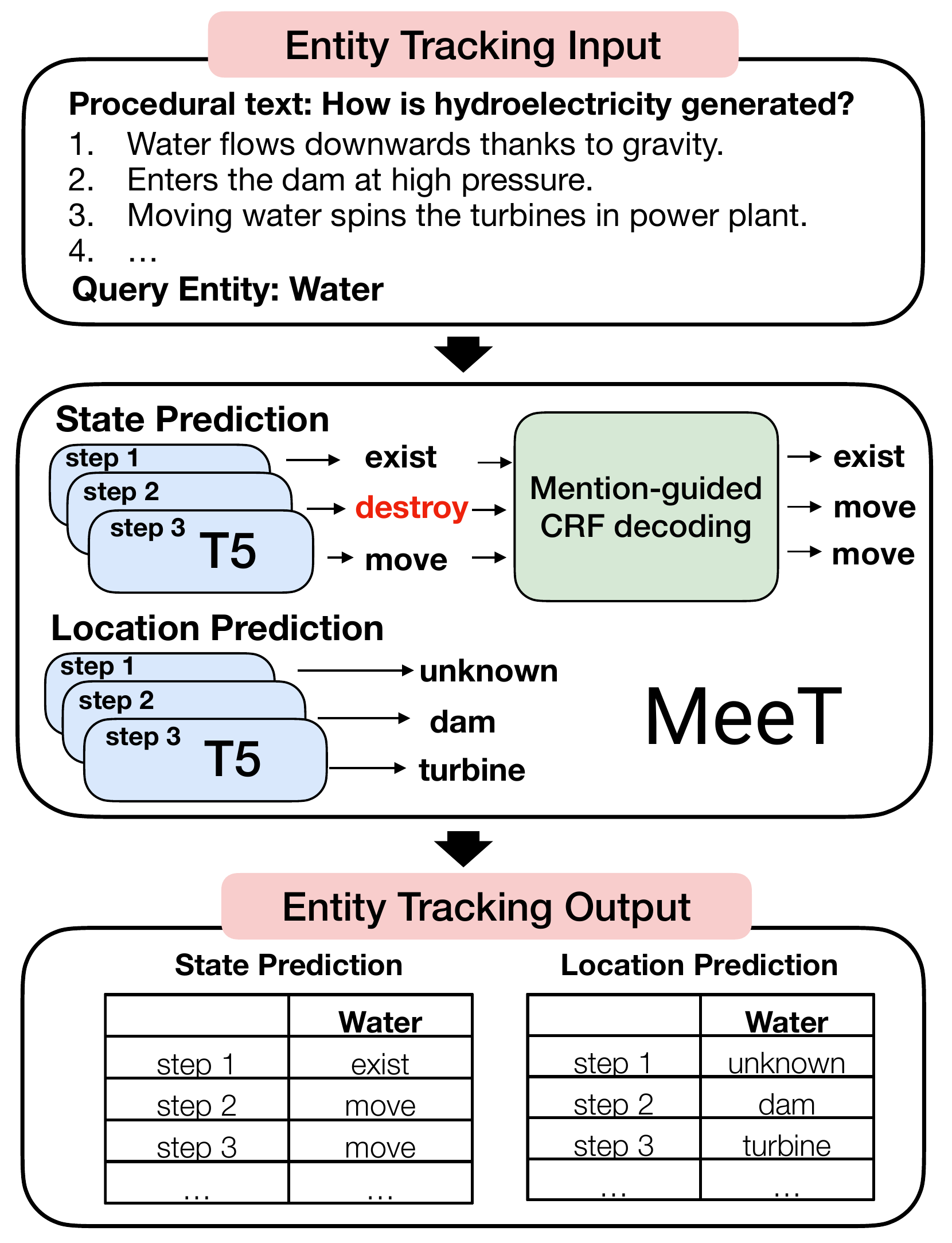}
\caption{Overview of \ours{} (\textbf{M}ulti-task learning-\textbf{e}nabled \textbf{e}ntity \textbf{T}racking). 
\ours{} utilizes the multi-task learning in T5 to boost entity tracking performance, with a customized decoding strategy addressing the structural constraints in state prediction (e.g., "move" cannot happen after "destroy").
  }
  \label{fig:main_figure}
\end{center}
\end{figure}

Pre-trained language models have revolutionized the NLP field in recent years \citep{devlin-etal-2019-bert, liu2019roberta, gpt3} and also become more versatile with the novel encoder-decoder architecture \citep{raffelT5, lewis-etal-2020-bart}, which allows them to handle different types of NLP tasks without further architectural changes. This versatility inherently facilitates cross-task knowledge transfer via multi-task learning \citep{raffelT5, aribandi2022ext}, and thus helps push the boundary of many popular NLP tasks such as question answering \citep{khashabi-etal-2020-unifiedqa} and semantic parsing \citep{UnifiedSKG}.
However, entity tracking, which tracks the states and locations of an entity throughout the procedural text, like scientific processes or recipes, has not been impacted by this multi-task learning wave for two main reasons.
First, entity tracking requires the model to make step-wise predictions while satisfying structural constraints (e.g., an entity cannot be "moved" after being "destroyed" in the previous steps). 
This requirement is usually tackled by designing task-specific architectures \citep{gupta-durrett-2019-tracking,tang-etal-2020-understanding-procedural,huang-etal-2021-reasoning}, and those generic multi-task models with the encoder-decoder architecture cannot address it easily.
Second, understanding procedural text requires domain-specific knowledge, which usually does not exist in general domain tasks that multi-task learning models are  trained on, so it is not clear how effective the knowledge transfer will be given this domain gap \citep{zhang_koala, bai-etal-2021-pre, lemon}.

In this paper, we study how entity tracking can benefit from the current multi-task learning paradigm and present \textbf{\ours{}}, a \textbf{M}ulti-task learning-\textbf{e}nabled \textbf{e}ntity \textbf{T}racking approach.
This approach includes two parts. The first part fine-tunes T5 \citep{raffelT5}, a model that has been pre-trained on a diverse set of NLP tasks and has shown great cross-task generalizability.
Here, we design entity tracking-specialized QA formats to accommodate the need to make step-specific predictions, while facilitating effective knowledge transfer from T5.
The second part resolves conflicted state predictions under structural constraints.
We use a customized offline CRF inference algorithm, where the main idea is to emphasize the predictions of steps, in which the query entity is explicitly mentioned, because the fine-tuned model performs better in those cases (Table \ref{tab:imp-exp-perf}).
On two benchmark datasets, ProPara \citep{dalvi-etal-2018-tracking} and Recipes \citep{Bosselut2017SimulatingAD}, our \ours{} outperforms previous state-of-the-art methods, which require extra domain-specific pre-training or data augmentation. We verify the importance of multi-task learning in T5 and our proposed decoding strategy through careful analyses and ablation studies.

To sum up, our contributions are three-fold: 
(1) Our work is the first to explore cross-task knowledge transfer for entity tracking on procedural text;
(2) Our proposed approach, \ours{}, effectively uses the off-the-shelf pre-trained multi-task learning model T5 with a customized decoding strategy, and thus achieves state-of-the-art performance on two benchmark datasets;
(3) Our comprehensive analyses verify the benefits of multi-task learning on entity tracking.

\section{Related Work}
\label{sec:related_work}
Tracking the progression of an entity within procedural text, such as cooking recipes \citep{Bosselut2017SimulatingAD} or scientific protocols \citep{tamari-etal-2021-process, le-etal-2022-shot, bai-etal-2022-synkb}, is challenging as it calls for a model to understand both superficial and intrinsic dynamics of the process. 
Recent work on entity tracking can be divided into two lines.
One focuses on designing task-specific fine-tuning architectures to ensure that the model makes step-grounded predictions while following the structural constraints.
For instance, \citet{rajaby-faghihi-kordjamshidi-2021-time} introduce time-stamp embeddings into RoBERTa \cite{liu2019roberta} to encode the index of the query step. 
\citet{gupta-durrett-2019-tracking} frame entity tracking as a structured prediction problem and use a CRF layer to promote global consistency under those structural constraints.
In our case, we show that, with QA formulation, simply appending the index of the query step to the question and indexing the procedure produces step-specific predictions.
Moreover, we propose a customized offline CRF-decoding strategy for structural constraints to compensate for the fact that it is hard to jointly train T5, our backbone LM, with a CRF layer, like in previous methods.

The other line of work focuses on domain-specific knowledge transfer \citep{zhang_koala, bai-etal-2021-pre, lemon, ma-etal-2022-coalescing}. 
Concretely, \lemon{} \cite{lemon} achieves great performance by performing in-domain pre-training on 1 million procedural paragraphs.
\cgli{} \citep{ma-etal-2022-coalescing} shows that adding high-quality pseudo-labeled data (generated via self-training) during fine-tuning can also boost the model performance. In contrast, our work explores how entity tracking can benefit from out-of-domain knowledge via using off-the-shelf pre-trained multi-task learning models.

\section{Method}
\label{sec:method}

In this section, we present \textbf{\ours{}}, a \textbf{M}ulti-task learning-\textbf{e}nabled \textbf{e}ntity \textbf{T}racking approach. Here, we first review the problem definition, and then lay out the details of \ours{}. 

\subsection{Problem Definition}
\label{sec:task_defnition}

Entity tracking aims at monitoring the status of an entity throughout a procedure.
The input of this task contains two items: 1) a procedural paragraph $P$, composed of a sequence of sentences $\{s_1, s_2, ..., s_T \}$; and 2) a procedure-specific query entity $e$.
Given the input, our goal is to predict the state and location of the query entity at each timestamp of the procedure (see an example from the ProPara dataset in Figure \ref{fig:main_figure}).

\subsection{\ours{}}
\ours{} includes two parts, task-specific fine-tuning with our proposed QA formats 
and the mention-guided conflict-resolve decoding.

\paragraph{Task-spefic Fine-tuning} We formulate the two sub-tasks of entity tracking, state prediction and location prediction, as multi-choice and extractive QA problems respectively (see \S\ref{sec:analysis} for comparison with other task formulations), and fine-tune T5 to make independent predictions for every step in the procedure. 
Given a query entity $e$ and procedure $P$, to predict the entity state at step $t$, the input sequence is formatted as the concatenation of the template question ``\textit{What is the state of $e$ in step $t$?}'', candidate states (e.g., \textit{create}, \textit{move} and \textit{destroy}), and the full procedure with step index prepended.
The output is just one of the candidate states.
For location prediction, the input sequence is the concatenation of the question ``\textit{Where is $e$ located in step $t$?}'' and the indexed procedure, with the snippet ``\textit{Other locations: none, unknown.}'' appended. This is because entity locations sometimes are not explicitly mentioned in the procedure.
The output is a text span, indicating the location of the query entity after step $t$. 
Examples of both tasks can be found in Appendix \ref{sec:format}.

\paragraph{Conflict-resolve Decoding} Entity tracking places unique structural constraints on state predictions (e.g.,  \textit{move} cannot happen after \textit{destroy}). Similar to \citet{gupta-durrett-2019-effective}, we run an offline CRF-decoding method (Viterbi decoding) to resolve conflicting state predictions. 
We initialize CRF transition scores $T$ with the transition statistics in the training data, following \citet{ma-etal-2022-coalescing}.
For example, $T(p, q)$, the transition score between state $p$ and $q$, is $log(1/10)$ if there is only one $p \Rightarrow q$ transition out of 10 transitions starting with the state $p$. We set the scores of all unseen transitions to $-inf$.
As for CRF emission scores, we use the state prediction logits from T5.
In contrast with previous methods, which treat each step equally, we weigh the emission scores differently, depending on whether the query entity $e$ is \textbf{explicitly mentioned} in the step:
\[
U_i^\prime = 
\begin{cases}
\tau_{exp} \cdot U_i, & \text{if } e \text{ is mentioned in step } i,\\
\tau_{imp} \cdot U_i, & \text{otherwise} \\
\end{cases}
\]
\normalsize
\noindent where $U_i^\prime$ represents the emission score of step $i$ after weighing, and $\tau_{exp}$ and $\tau_{imp}$ are hyper-parameters, determined by the grid search on the dev set.
The intuition behind our approach is that, as the fine-tuned model performs better on "explicitly mentioned" steps (Table \ref{tab:imp-exp-perf}), leaning toward those steps during decoding via controlled weights will result in more accurate predictions.\footnote{After hyper-parameter tuning, the optimal values for $\tau_{exp}$ and $\tau_{imp}$ are 0.6 and 0.7 respectively.}

\begin{table}[t!]
\begin{center}
\setlength{\tabcolsep}{5pt}
\scalebox{0.9}{
\begin{tabular}{lccc}
\toprule


\textbf{Model}
& \bf{P}
& \bf{R}
& \bf{F1}

\\

\midrule




\dynapro{} \citep{amini2020procedural} 
& 75.2 
& 58.0 
& 65.5

\\ 

\tslm{}\textsuperscript{\textdagger} \citetalias{rajaby-faghihi-kordjamshidi-2021-time}
& 68.4 
& 68.9
& 68.6

\\ 
 
\koala\textsuperscript{\textdagger} \citep{zhang_koala}
& 77.7
& 64.4 
& 70.4

 \\ 



\lemon{}\textsuperscript{\textdagger}\citep{lemon}
& 74.8 
& 69.8
& 72.2
\\ 




\cgli{}\textsuperscript{\textdagger} \citep{ma-etal-2022-coalescing}
& 75.7
& \textbf{70.0}
& 72.7

\\

\midrule


\textbf{\ours{}} (ours) 


& \textbf{80.3} & 67.1 & \textbf{73.1}

 \\

\bottomrule 
\end{tabular}
 }
\end{center}

\caption{\label{tab:result_propara_merged} Test set performance on ProPara.
\textsuperscript{\textdagger} indicates that the backbone language model has been further pre-trained on either in-domain corpus or auxiliary tasks. \ours{} performs on par with SOTA models without pre-finetuning on any in-domain corpus.} 

\end{table}

\section{Experiments}
\label{sec:exp}
\paragraph{Datasets} 
We experiment with two benchmark datasets of entity tracking: ProPara \citep{dalvi-etal-2018-tracking} and Recipes \cite{Bosselut2017SimulatingAD}. 
ProPara contains 488 scientific process-based procedural paragraphs (Figure \ref{fig:main_figure}), and Recipes includes 866 cooking recipes.
Note that previous work experiments with different splits of the Recipes dataset; in this paper, we follow the split of \citet{zhang_koala}\footnote{ {\url{https://github.com/ytyz1307zzh/KOALA/issues/4}}} as it is used in most of the recent work \citep{huang-etal-2021-reasoning, lemon}.
More dataset details are presented in Appendix~\ref{sec:appdx_dataset}.

\paragraph{Evaluation} 
ProPara performances are evaluated in two levels: \textit{sentence-level}\footnote{{\url{https://github.com/Mayer123/CGLI/blob/main/src/evalQA.py}}} \citep{dalvi-etal-2018-tracking} and \textit{document-level}\footnote{{\url{https://github.com/allenai/aristo-leaderboard/blob/master/propara/evaluator}}} \citep{tandon-etal-2018-reasoning}.
Here, we focus on the \textit{document-level} evaluation because it provides a comprehensive assessment of the model's understanding of the overall procedure and serves as the basis for the ProPara leaderboard rankings.
The \textit{document-level} evaluation is conducted by comparing the input/output entities and their transformations in the procedure with the gold answers.
Further details regarding two evaluations and the result of the \textit{sentence-level} evaluation can be found in Appendix \ref{sec:eval_details_appdx}.
For Recipes, following previous work \citep{zhang_koala, lemon}, we evaluate the location changes of each ingredient throughout the recipe.\footnote{ {\url{https://drive.google.com/drive/folders/1PYGLe7hSoCYfpKmpPumeTy6jmPyONGz4}}}

\paragraph{Baselines} For ProPara, we compare \ours{} with the top five approaches on its leaderboard. 
Among these five approaches, \dynapro{} \citep{amini2020procedural}, \tslm{} \citep{rajaby-faghihi-kordjamshidi-2021-time}, and \cgli{} \citep{ma-etal-2022-coalescing} design task-specific fine-tuning architecture using off-the-shelf LMs while \koala{} \citep{zhang_koala} and \lemon{} \citep{lemon} develop in-domain LMs for procedural text.
For Recipes, as mentioned previously, we compare \ours{} with methods that experiment on the same data split of \citet{zhang_koala}.
We refer readers to the corresponding paper of each baseline for further details.

\paragraph{Implementation Details} Our approach \ours{} is implemented using Huggingface Transformers \citep{wolf-etal-2020-transformers}. Given the limited computational resources, we choose T5-large as the backbone of our \ours{}. The fine-tuning process employs the AdamW optimizer with a learning rate of $1 \times 10^{-4}$ and a batch size of 16.
To resolve any potential conflict between state prediction and location prediction, we apply the rules designed in \citet{ma-etal-2022-coalescing} to integrate the output from both tasks.


\subsection{Results}
\label{sec:results}

We present the test set results of ProPara and Recipes in Table \ref{tab:result_propara_merged} and Table \ref{tab:result_recipe}, respectively.
Our \ours{} outperforms the competitive baseline \cgli{} \citep{ma-etal-2022-coalescing} on the ProPara dataset with state-of-the-art performance despite the fact that \cgli{} uses extra pseudo-labeled training data (generated by self-training) for data augmentation. 
On Recipes, \ours{} surpasses the previous best-performing method \lemon{} \citep{lemon} by a substantial margin of 4.9 F\textsubscript{1}. 
It is noteworthy that the cooking recipes in the Recipes dataset were collected from the web,\footnote{\url{http://www.ffts.com/recipes.htm}} which may have been included in the C4 corpus\footnote{\url{https://www.tensorflow.org/datasets/catalog/c4}} used for pre-training T5 and thus potentially contributes to the advantage of our \ours{} on Recipes.

\subsection{Analysis \& Ablation Study}
\label{sec:analysis}

\paragraph{Multi-task Learning} To investigate the impact of T5's multi-task learning process on entity tracking, we experiment with two variants of T5 as the backbone of \ours{}: 1) T5-v1.1,\footnote{\url{https://huggingface.co/docs/transformers/model_doc/t5v1.1}} a T5-like LM (with slight architecture changes) whose pre-training does not include any supervised tasks; 2) T5-v1.1\textsubscript{\textit{QA-FT}}, the resulting LM after fine-tuning T5-v1.1 on the three QA datasets,\footnote{The three datasets include MultiRC \citep{khashabi-etal-2018-looking}, ReCoRD \citep{Zhang2018ReCoRDBT}, and BoolQ \citep{clark-etal-2019-boolq}} which T5 is pre-trained on.
The performance of the three LMs (T5-large size) on the ProPara dev set is presented in the top section of Table \ref{tab:ablation}.
We can see that T5 outperforms T5-v1.1 by a large margin, verifying that multi-task learning on out-of-domain non-entity-tracking tasks can benefit entity tracking.
In addition, the advantage of T5 over T5-v1.1\textsubscript{\textit{QA-FT}} indicates that knowledge transfer can cross the task boundaries with T5's encoder-decoder architecture.

\begin{table}[t!]
\begin{center}
\setlength{\tabcolsep}{5pt}
\scalebox{0.85}{
\begin{tabular}{lccc}

\toprule

\textbf{Model}
& \textbf{P}
& \textbf{R}
& \textbf{F\textsubscript{1}} \\

\midrule

\ncet{} \citep{gupta-durrett-2019-tracking} 
& 56.5
& 46.4	
& 50.9\\ 

\ien{} \citep{tang-etal-2020-understanding-procedural} 
& 58.5
& 47.0	
& 52.2\\ 
 
\koala{} \citep{zhang_koala}
& 60.1
& 52.6
& 56.1\\ 

\real{} \citep{huang-etal-2021-reasoning}
& 55.2
& 52.9
& 54.1\\ 


\lemon{} \citep{lemon}
& 56.0
& 67.1
& 61.1\\ 



\midrule






\textbf{\ours{}} (ours)
& \textbf{64.2} & \textbf{78.0} & \textbf{66.0} \\

\bottomrule 
\end{tabular}
}
\end{center}

\caption{\label{tab:result_recipe} Test set results on Recipes. \ours{} achieves the state-of-the-art performance, outperforming the previous SOTA \lemon{} by 4.9 F\textsubscript{1}.}

\end{table}

\begin{table}[t!]
\begin{center}
\setlength{\tabcolsep}{5pt}
\scalebox{0.8}{
\begin{tabular}{lccc}

\toprule

&\textbf{P}
& \textbf{R}
& \textbf{F1} \\

\midrule

\textbf{\ours{}} (ours) & 77.3 & 71.1 &	74.1   \\

\midrule

\multicolumn{ 4}{c}{\textit{Multi-task Learning}} \\

\midrule

T5-v1.1 &	76.6 &	64.9 &	70.3 \\

T5-v1.1\textsubscript{\textit{QA-FT}} &	76.3 &	65.8 &	70.7 \\

\midrule

\multicolumn{ 4}{c}{\textit{Task Formulation }} \\

\midrule

Process-level & 89.3 & 30.2 &	45.1  \\
Step-level &	76.7 &	61.3 &	68.1  \\

\midrule

\multicolumn{ 4}{c}{\textit{Decoding Strategy \& Model Size}} \\

\midrule
CRF-normal & 75.0 &	72.8 &	73.8  \\

T5-base & 76.8 & 68.2 &	72.2  \\

\bottomrule 
\end{tabular}
}
\end{center}

\caption{\label{tab:ablation} 
Analysis and ablation study on ProPara (dev set results). Top: Comparison of different backbone LMs to investigate the impact of multi-task learning. Middle: Comparison of different task formulations.
Bottom: Ablation on decoding strategy and model size.
Multi-task learning leads to a better entity tracking model, especially with the QA formulation and mention-guided decoding. 
}

\end{table}

\paragraph{Task Formulation}
We compare our QA formulation with two other task formulations, proposed in recent work, for T5. 
The first formulation is called "\textit{step-input}" \citep{gupta-durrett-2019-effective,amini2020procedural}, where each pair of the query entity $e$ and procedure step $t$ is formulated as one instance.
Here, the state prediction is formulated as a classification problem, where the entity name is appended to the input, and no candidate answers are provided. Moreover, the procedure is trimmed until step $t$ to specify the step index in the input.
The second formulation is called "\textit{process-input}" \citep{zhang_koala, gupta-durrett-2019-tracking}, where the model predicts entity states or locations in all steps in one instance.
The input is the concatenation of entity $e$ and the full procedure, and the model decodes entity states and locations in all steps sequentially.
The results of two new formulations are presented in the middle of Table \ref{tab:ablation}.
Our proposed QA formulation outperforms the other two formulations by a large margin.
Detailed analyses of formulation comparison can be found in Appendix \ref{sec:form_comp}.

\paragraph{Decoding Strategy \& Model Size} The ablation study on decoding strategy and model size is shown at the bottom section of Table \ref{tab:ablation}. Clearly, our proposed "mention-guided" decoding strategy, as well as using a larger LM as the backbone, contribute to the success of \ours{}.

\section{Conclusion}
\label{sec:conclu}

We presented \ours{}, a T5-based entity tracking approach. This approach includes our newly proposed QA fine-tuning formats and a customized decoding strategy so that it can effectively encode the flow of events in the procedural text while following structural constraints. The state-of-the-art performances on two benchmark datasets demonstrate the effectiveness of \ours{}, and further analyses verify that multi-task learning on out-of-domain tasks can be beneficial for entity tracking.

\section*{Limitations}
\label{sec:limit}

This paper demonstrates that multi-task learning on a combination of general domain datasets can effectively improve the model's understanding of the procedural text. However, the precise source dataset responsible for this improvement remains uncertain, making it an avenue for future research to investigate more efficient knowledge transfer through the identification of the most pertinent source dataset.
Moreover, the pipeline structure of \ours{} may limit its practical utilization. As such, future work could consider incorporating our proposed mention-guided decoding strategy into the end-to-end training of the multi-task learning model.


\bibliography{anthology,custom}
\bibliographystyle{acl_natbib}

\clearpage

\appendix

\label{sec:appendix}

\section{Fine-tuning Formats for T5}
\label{sec:format}
\subsection{State Prediction (Multi-choice QA)}
Input:
\begin{displayquote}
\small
\texttt{What is the state of water in step 2? \\}
\texttt{(a) create (b) ... (f) move\\}
\texttt{step 1: Water flows downawards thanks to gravity. step 2: Enters the dam at high pressure. step 3: The moving water spins the
turbines in the power plant ... step 6:
The water leaves the dam at the bottom.}
\end{displayquote}

\noindent Output:
\begin{displayquote}
\small
\texttt{move\\}
\end{displayquote}

\subsection{Location Prediction (Extractive QA)}
Input:
\begin{displayquote}
\small
\texttt{Where is water located in step 2? \\}
\texttt{step 1: Water flows downwards thanks to gravity. step 2: Enters the dam at high pressure. step 3: The moving water spins the
turbines in the power plant ... step 6:
The water leaves the dam at the bottom. Other locations: none, unknown.}
\end{displayquote}

\noindent Output:
\begin{displayquote}
\small
\texttt{dam\\}
\end{displayquote}

\section{Dataset}
\label{sec:appdx_dataset}
\begin{table}[t!]
\begin{center}
\setlength{\tabcolsep}{5pt}
\scalebox{0.8}{
\begin{tabular}{ll|ccc|c}

\toprule

\textbf{Dataset} & \textbf{Statistics} & \textbf{Train} & \textbf{Dev} & \textbf{Test} & \textbf{Total}\\

\midrule 

\multirow{3}{*}{\textbf{Recipes}} 
&\# procedures & 693 & 86 & 87 & 866\\
& Avg. steps / proc. & 8.8 & 8.9 & 9.0 & 8.8 \\ 
& Avg. entities / proc. & 8.6 & 8.8 & 8.5 & 8.6 \\

\midrule 

\multirow{ 3}{*}{\textbf{ProPara}} 
&\# procedures & 391 & 43 & 54 & 488 \\
& Avg. steps / proc. & 6.8 & 6.7 & 6.9 & 6.8 \\ 
& Avg. entities / proc. & 3.8 & 4.1 & 4.4 & 3.9 \\

\bottomrule 
\end{tabular}
}
\end{center}

\caption{\label{tab:data_stat}Statistics of Recipes and ProPara.}
\end{table}
\begin{table}[t!]
\begin{center}
\setlength{\tabcolsep}{5pt}
\scalebox{0.8}{
\begin{tabular}{lccc}
\toprule
& \textbf{P}
& \textbf{R} 
& \textbf{F1} \\
\midrule
\textit{implicit} &	 37.4  & 23.2  & 28.3 \\
\textit{explicit} &	 \textbf{68.3}  & \textbf{72.4}  & \textbf{70.2} \\

\bottomrule 
\end{tabular}
}
\end{center}
\caption{\label{tab:imp-exp-perf}  
\ours{}'s sentence-level performance (before applying offline CRF) on \textit{implicit} and \textit{explicit} steps (where the query entity is explicitly mentioned). Clearly, \ours{} makes more accurate predictions on \textit{explicit} steps.
}

\end{table}

For ProPara \citep{dalvi-etal-2018-tracking}, following \citet{ma-etal-2022-coalescing}, the state prediction task includes six candidate states (\texttt{Outside\_Before}, \texttt{Create}, \texttt{Destroy}, \texttt{Move}, \texttt{Exist} and \texttt{Outside\_After}). 
For Recipes \cite{Bosselut2017SimulatingAD}, each ingredient has two possibles states (\texttt{Exist} or \texttt{Absence}) in each step of the recipe.
Full data statistics on two datasets are presented in Table \ref{tab:data_stat}.

\section{Evaluation}
\label{sec:eval_details_appdx}

\paragraph{Sentence-level evaluation} This evaluation measures the following questions for each target \textit{entity}: 
\begin{itemize}
\item \textbf{Cat-1:} Is \textit{entity} created (destroyed, moved) in the process?
\item \textbf{Cat-2:} When (step \#) is \textit{entity} created (destroyed, moved)?
\item \textbf{Cat-3:} Where (location) is \textit{entity} created (destroyed, moved to/from)?
\end{itemize}
Further, the F\textsubscript{1} scores of the three questions are aggregated with micro/macro averages.

\paragraph{Document-level evaluation} It measures the four questions below for each paragraph: 

\begin{itemize}
\item What are the \textit{input} entities to the process?
\item What are the \textit{output} entities of the process?
\item What entity \textit{conversions} occur, when (step \#), and where (location)?
\item What entity \textit{movements} occur, when, and where?
\end{itemize}
The macro average of the F\textsubscript{1} scores of these  four questions will be used as the final score. 

\begin{table*}[t!]
\begin{center}
\setlength{\tabcolsep}{5pt}
\begin{tabular}{lcccccccc}
\toprule

\multirow{2}{*}{ \vspace{-0.3cm} \textbf{Model}}
& \multicolumn{3}{c}{\textbf{Document-level} }
& \multicolumn{5}{c}{\textbf{Sentence-level} }
\\

\cmidrule(lr){2-4} \cmidrule(lr){5-9}

& \bf{P}
& \bf{R}
& \bf{F1}

& \textbf{Cat-1}
& \textbf{Cat-2}
& \textbf{Cat-3} 
& \bf{macro}
& \bf{micro} 
\\

\midrule



\ncet{} \citep{gupta-durrett-2019-tracking} 

& 67.1
& 58.5
& 62.5

& 73.7
& 47.1
& 41.0
& 53.9
& 54.0 \\

IEN \cite{tang-etal-2020-understanding-procedural}  & 69.8  & 56.3  & 62.3  & 71.8  & 47.6 &  40.5  & 53.3 &  53.0 \\

\dynapro{} \citep{amini2020procedural} 

& 75.2 
& 58.0 
& 65.5

& 72.4
& 49.3
& 44.5
& 55.4 
& 55.5
\\ 

ProGraph \cite{DBLP:journals/corr/abs-2004-12057}  & 67.3  & 55.8 &  61.0  & 67.8 &  44.6 &  41.8  & 51.4  & 51.5 \\

\tslm{} \citetalias{rajaby-faghihi-kordjamshidi-2021-time}

& 68.4 
& 68.9
& 68.6 

& 78.8
& 56.8	
& 40.9
& 58.8
& 58.4
\\ 
 
\koala{} \citep{zhang_koala}

& 77.7
& 64.4 
& 70.4

& 78.5
& 53.3
& 41.3
& 57.7
& 57.5
 \\ 
 
REAL \cite{huang-etal-2021-reasoning} &  \textbf{81.9}  & 61.9  & 70.5  & 78.4  & 53.7 &  42.4  & 58.2  & 57.9 \\




\lemon{} \citep{lemon}
& 74.8 
& 69.8
& 72.2

& \textbf{81.7}
& 58.3
& 43.3 
& 61.1
& 60.7
\\ 

\cgli{} \citep{ma-etal-2022-coalescing}
& 75.7
& \textbf{70.0}
& 72.7

& 80.8
& 60.7
& 46.8
& \textbf{62.8}
& \textbf{62.4} \\

\midrule



  




 \textbf{\ours{}} (ours)
 & 80.3 & 67.1 & \textbf{73.1}
&  77.5	 & \textbf{61.0} & \textbf{49.6} & 62.7 & \textbf{62.4} \\


\bottomrule 
\end{tabular}
\end{center}
\caption{\label{tab:result_propara_merged_appendix} Document-level and sentence-level evaluation results on ProPara test set. 
} 
\end{table*} 

Table \ref{tab:result_propara_merged_appendix} provides a comprehensive comparison of past work on the ProPara dataset, including both document-level and sentence-level evaluations.

\section{Analysis of Formulation Comparison}
\label{sec:form_comp}
When compared with the "\textit{step-input}" formulation, the QA formulation allows the model to have the full context, and may take better advantage of LM's pre-training scheme \citep{li-etal-2019-entity, nagata-etal-2020-supervised}.
The "\textit{process-input}" formulation works the worst in this comparison. With qualitative analyses, we find that it suffers from error propagation due to its autoregressive decoding, so future work may explore incorporating structural decoding \citep{tandon-etal-2018-reasoning} into T5.

\end{document}